\documentclass{ecai}
\usepackage{times}
\usepackage{graphicx}
\usepackage{latexsym}

\usepackage{amsmath}
\usepackage{amssymb}
\usepackage{nicefrac}
\usepackage{subcaption}
\newtheorem{definition}{Definition}

\usepackage{algorithm, algpseudocode}

\algblock{ParFor}{EndParFor}
\algnewcommand\algorithmicparfor{\textbf{for}}
\algnewcommand\algorithmicpardo{\textbf{do in parallel}}
\algnewcommand\algorithmicendparfor{\textbf{end\ for}}
\algrenewtext{ParFor}[1]{\algorithmicparfor\ #1\ \algorithmicpardo}
\algrenewtext{EndParFor}{\algorithmicendparfor}

\usepackage{array}

\usepackage{color}

\DeclareMathOperator{\comb}{\gamma}
\DeclareMathOperator{\aggr}{\textsc{Aggr}}
\DeclareMathOperator{\mess}{\phi}
\DeclareMathOperator{\saggr}{\textsc{Aggr}}
\DeclareMathOperator{\NN}{\mathcal{N}}

\DeclareMathOperator{\OO}{\mathcal{O}}
\DeclareMathOperator{\RR}{\mathbb{R}}
\DeclareMathOperator{\Akipf}{\hat{\tilde{A}}}
\DeclareMathOperator{\concat}{\scalebox{1}[1.5]{$\parallel$}}
\newcommand{\nz}[1]{\text{nz} \left( #1 \right)}
\newcommand{\nnz}[1]{\text{nnz} \left( #1 \right)}

\newcommand{\citet}{\cite}
\newcommand{\citep}{\cite}
\usepackage{url}
\usepackage{booktabs} 
\usepackage{tabularx}
\usepackage{balance}
\usepackage{cleveref}
\usepackage{placeins}
\usepackage{enumitem}

\begin{document}

\title{PushNet: Efficient and Adaptive Neural Message Passing}

\author{Julian Busch \institute{Siemens Corporate Technology, Princeton, NJ, USA \newline jiaxing.pi@siemens.com} \institute{Ludwig-Maximilians-Universit\"at M\"unchen, Munich, Germany \newline \{busch, seidl\}@dbs.ifi.lmu.de} \and Jiaxing Pi \footnotemark[1] \and Thomas Seidl \footnotemark[2] }

\maketitle
\bibliographystyle{ecai}

\begin{abstract}
    Message passing neural networks have recently evolved into a state-of-the-art approach to representation learning on graphs. Existing methods perform synchronous message passing along all edges in multiple subsequent rounds and consequently suffer from various shortcomings: Propagation schemes are inflexible since they are restricted to $k$-hop neighborhoods and insensitive to actual demands of information propagation. Further, long-range dependencies cannot be modeled adequately and learned representations are based on correlations of fixed locality. These issues prevent existing methods from reaching their full potential in terms of prediction performance.
Instead, we consider a novel asynchronous message passing approach where information is pushed only along the most relevant edges until convergence. Our proposed algorithm can equivalently be formulated as a single synchronous message passing iteration using a suitable neighborhood function, thus sharing the advantages of existing methods while addressing their central issues. The resulting neural network utilizes a node-adaptive receptive field derived from meaningful sparse node neighborhoods. In addition, by learning and combining node representations over differently sized neighborhoods, our model is able to capture correlations on multiple scales. We further propose variants of our base model with different inductive bias.
Empirical results are provided for semi-supervised node classification on five real-world datasets following a rigorous evaluation protocol. We find that our models outperform competitors on all datasets in terms of accuracy with statistical significance. In some cases, our models additionally provide faster runtime.
\end{abstract}

\section{Introduction}
As a natural abstraction of real-world entities and their relationships, graphs are widely adopted as a tool for modeling machine learning tasks on relational data. Applications are manifold, including documents classification in citation networks, user recommendations in social networks or function prediction of proteins in biological networks.

\begin{figure}[t]
	\centering
	\begin{subfigure}[b]{0.15\textwidth}
		\centering
		\includegraphics[width=\textwidth]{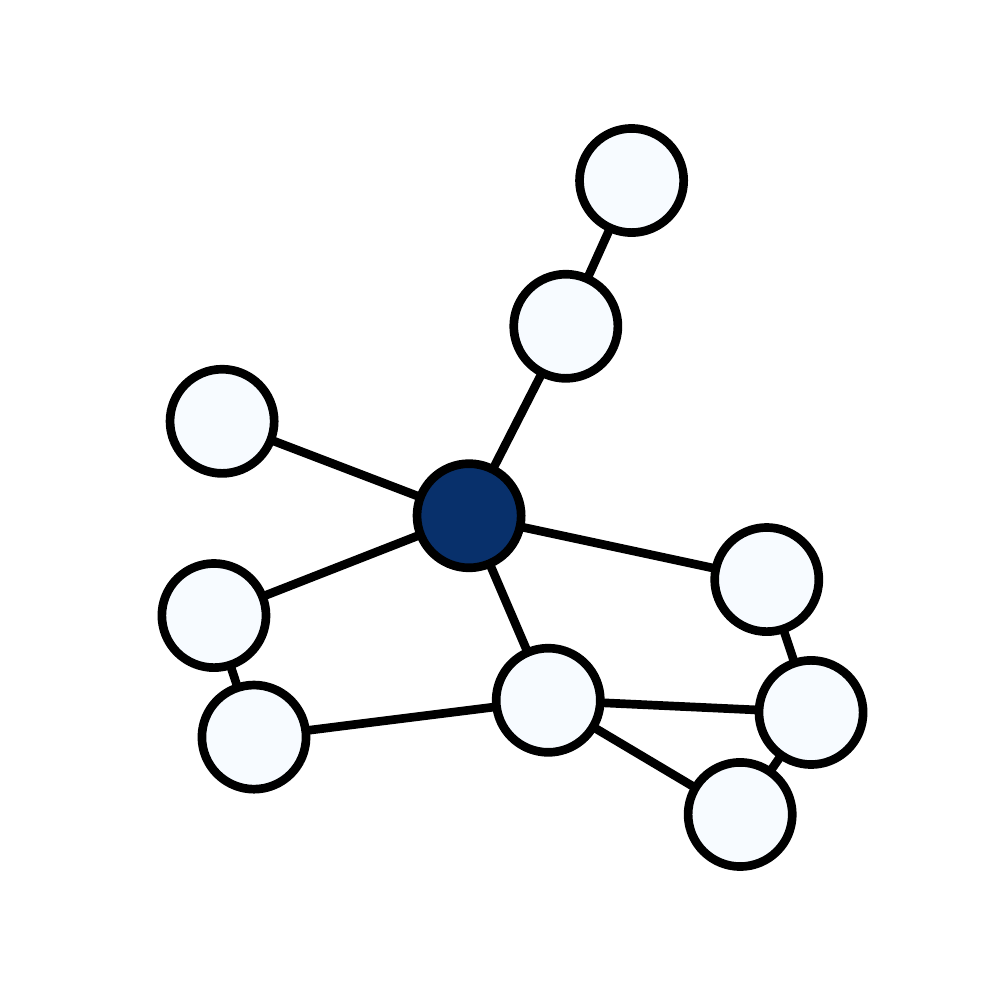}
		\caption{}\label{fig:diff_first}
	\end{subfigure}
	\begin{subfigure}[b]{0.15\textwidth}
		\centering
		\includegraphics[width=\textwidth]{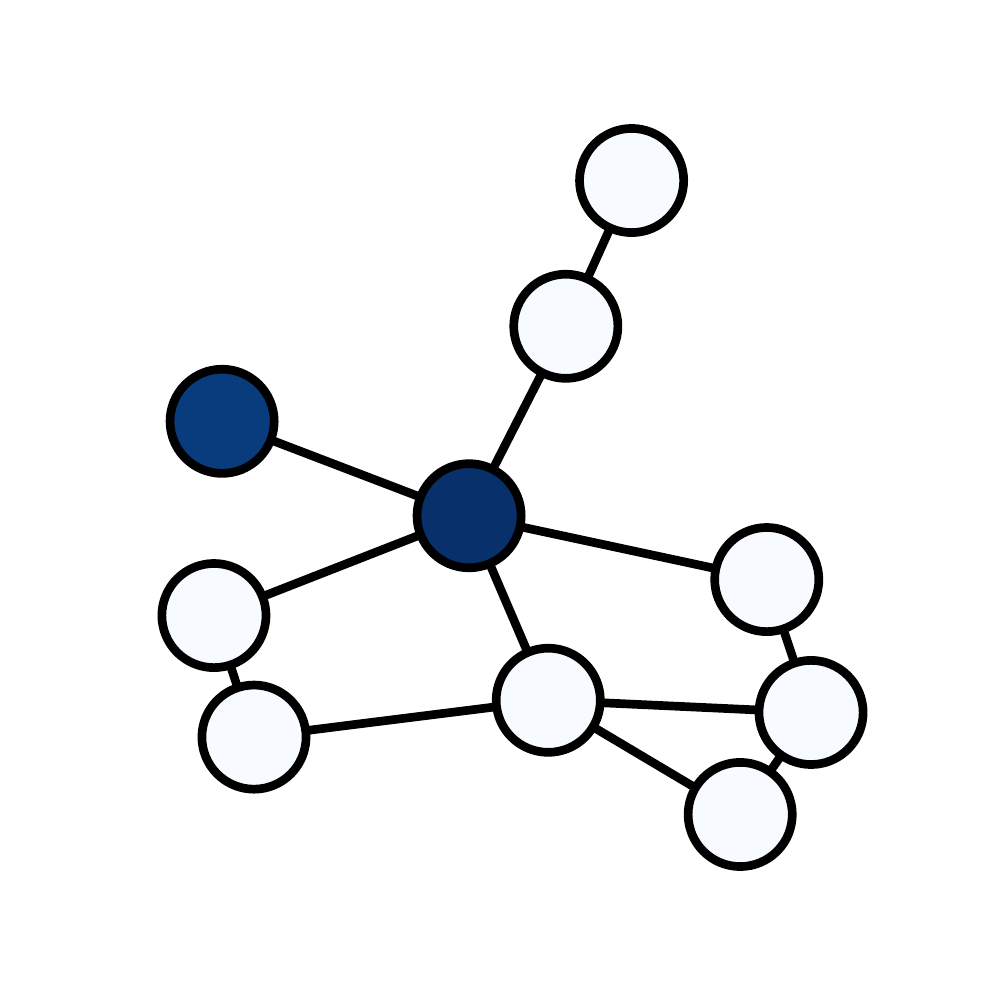}
		\caption{}
	\end{subfigure}
	\begin{subfigure}[b]{0.15\textwidth}
		\centering
		\includegraphics[width=\textwidth]{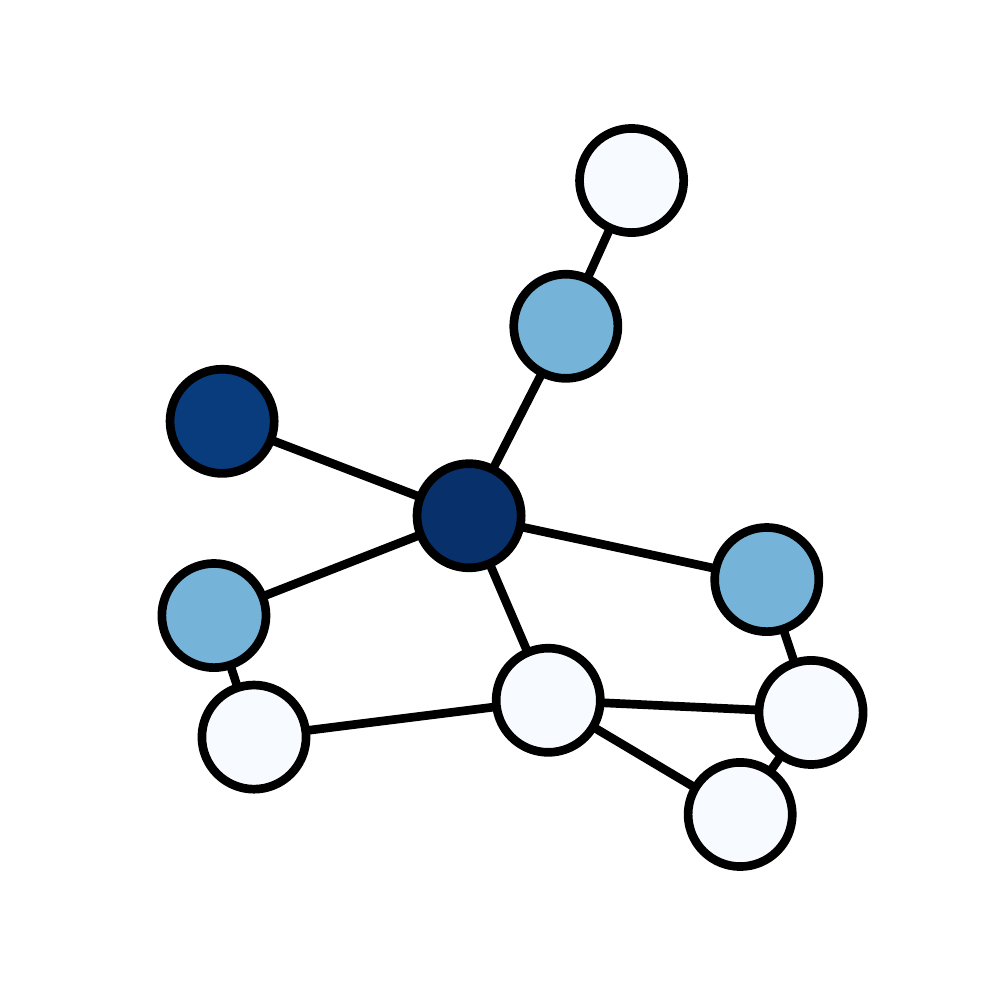}
		\caption{}
	\end{subfigure}

	\begin{subfigure}[b]{0.15\textwidth}
		\centering
		\includegraphics[width=\textwidth]{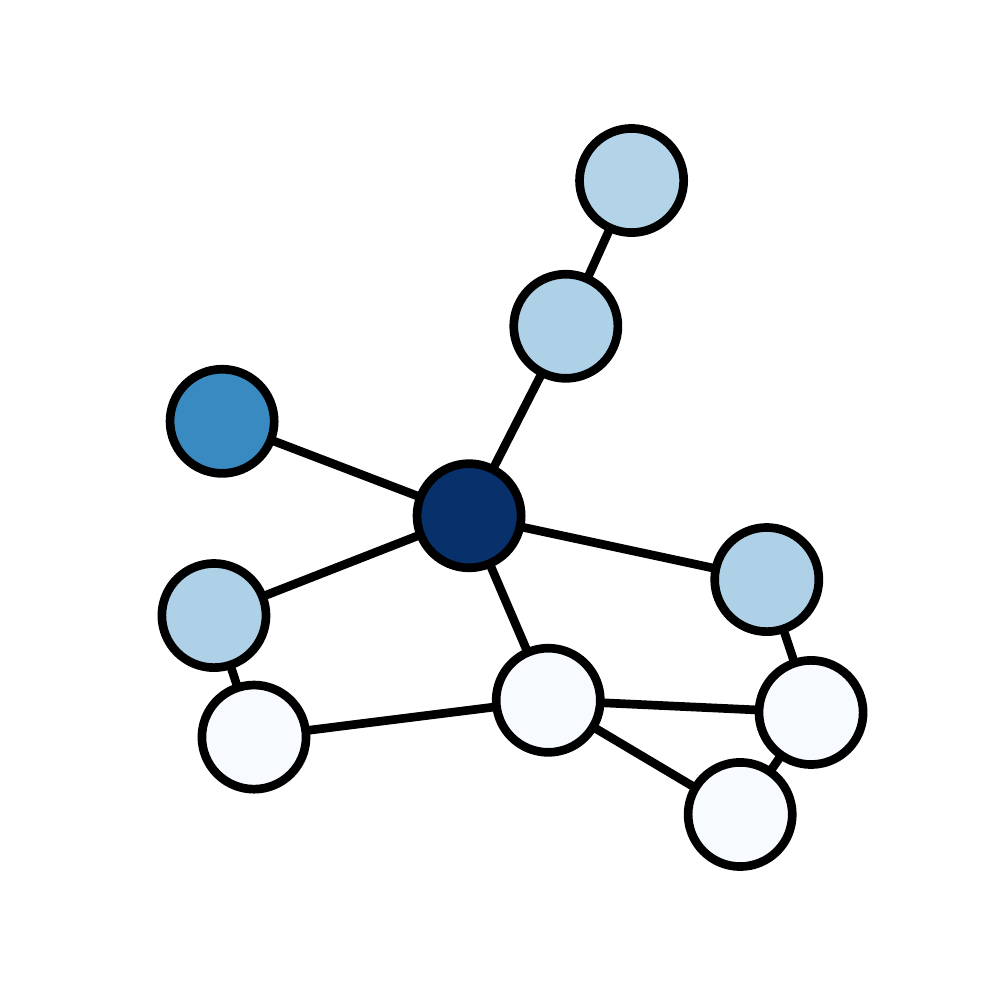}
		\caption{}
	\end{subfigure}
	\begin{subfigure}[b]{0.15\textwidth}
		\centering
		\includegraphics[width=\textwidth]{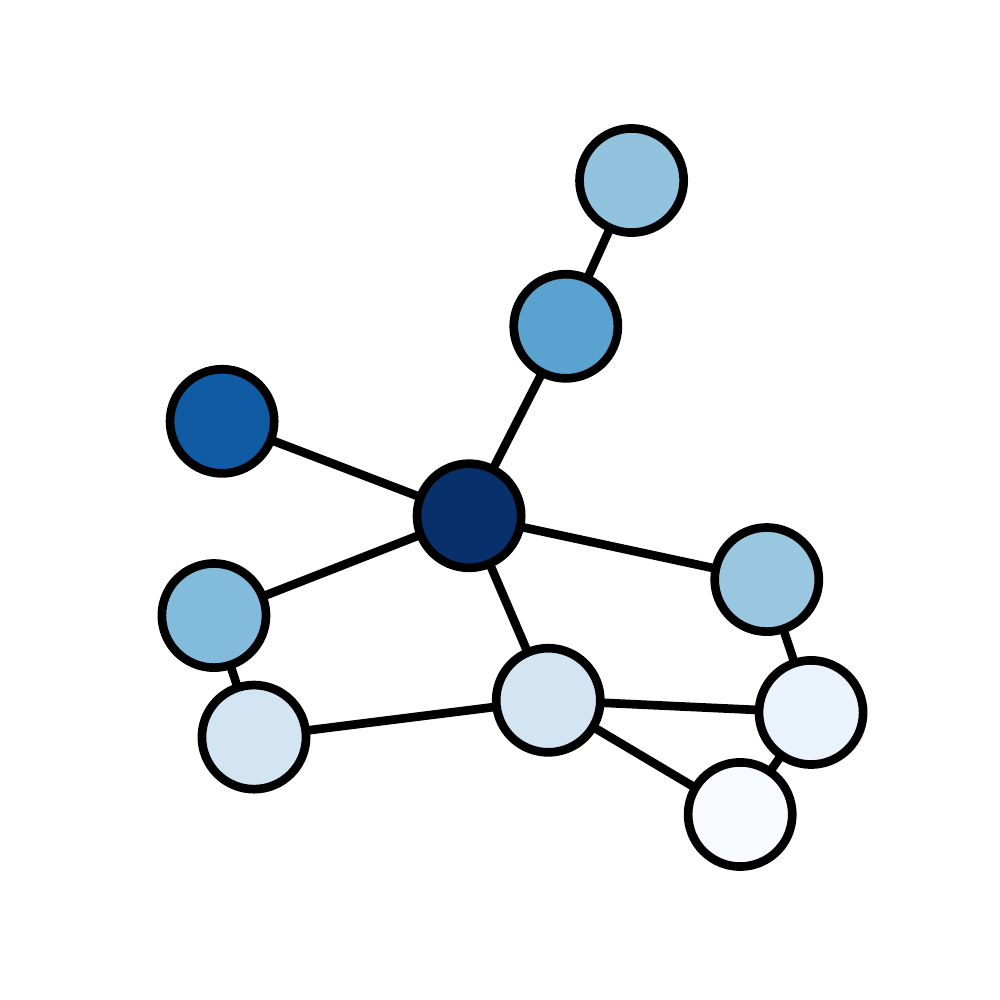}
		\caption{}\label{fig:diff_last}
	\end{subfigure}
	\begin{subfigure}[b]{0.15\textwidth}
		\centering
		\includegraphics[width=\textwidth]{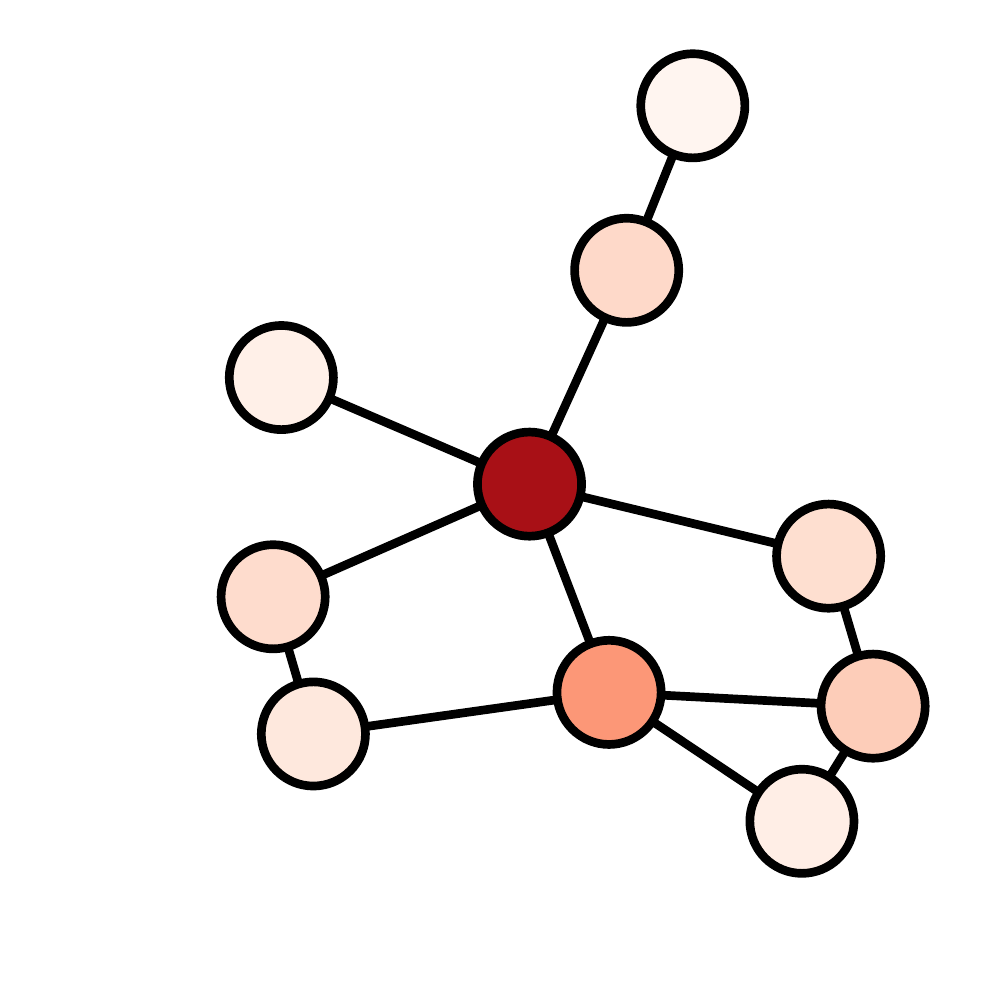}
		\caption{}\label{fig:diff_imp}
	\end{subfigure}
	
	\caption{Features of the central node are pushed through the graph until convergence (\ref{fig:diff_first}--\ref{fig:diff_last}). Equivalently, each node performs a single aggregation step over a node-adaptive neighborhood with importance weights shown for the central node in (\ref{fig:diff_imp}).}
	\label{fig:example}
\end{figure}

Remarkable success has been achieved by recent efforts in formulating deep learning models operating on graph-structured domains. Unsupervised node embedding techniques \citep{perozzi2014deepwalk,tang2015line,cao2015grarep,grover2016node2vec,qiu2018network} employ matrix factorization to derive distributed vector space representations for further downstream tasks. In settings where labels are provided, semi-supervised models can be trained end-to-end to improve performance for a given task. In particular, graph neural network models \citep{bronstein2017geometric,gilmer2017neural,battaglia2018relational} have been established as a de-facto standard for semi-supervised learning on graphs. While spectral methods \citep{bruna2013spectral,defferrard2016convolutional,monti2017geometric,bronstein2017geometric} can be derived from a signal processing point of view, a message passing perspective \citep{duvenaud2015convolutional,li2015gated,kearnes2016molecular,kipf2016semi,hamilton2017inductive,gilmer2017neural} has proved especially useful due to its flexibility and amenability to highly parallel GPU computation \citep{fey2019fast}. 
Further recent works have considered additional edge features \citep{gilmer2017neural,schlichtkrull2018modeling}, attention mechanisms \citep{velivckovic2017graph,thekumparampil2018attention,lee2018attention}, addressed scalability \citep{chen2018fastgcn,wu2019simplifying} and studied the expressive power of graph neural network models \citep{xu2018powerful,morris2018weisfeiler}.

While the above techniques may serve as a basis for modeling further tasks such as link prediction \citep{kipf2016variational,zhang2018link} or graph classification \citep{niepert2016learning,lee2018graph}, we focus on \emph{semi-supervised node classification} in this work. Given a graph $G = (V, E)$, a feature matrix $X \in \RR^{n \times d}$ and a label matrix $Y \in \RR^{n \times c}$, the goal is to predict labels for a set of unlabeled nodes based on graph topology, node attributes and observed node labels. If no node attributes are available, auxiliary features such as one-hot vectors or node degrees may be used, depending on the task at hand. 
All graphs considered in the following are undirected, however, extension to directed graphs is straightforward.

Despite their success, existing neural message passing algorithms suffer from several central issues. First, information is pulled indiscriminately from $k$-hop neighborhoods which will include many irrelevant nodes and miss important ones. In particular, long-range dependencies are modeled ineffectively, since unnecessary messages do not only impede efficiency but additionally introduce noise. Further, interesting correlations might exist on different levels of locality which makes it necessary to consider multi-scale representations. These issues prevent existing neural message passing algorithms from reaching their full potential in terms of prediction performance. 

To address the above issues, we propose a novel $push$-based neural message passing algorithm which propagates information on demand rather than indiscriminately pulling it from all neighbors. We show that it can be interpreted equivalently as either an asynchronous message passing scheme or a single synchronous message passing iteration over sparse neighborhoods derived from \emph{Approximate Personalized PageRank}. Thereby, each node neighborhood is personalized to its source node, providing a stronger structural bias and resulting in a node-adaptive receptive field. Both views are illustrated in Figure \ref{fig:example}. Consequently, our model benefits from the existing synchronous neural message passing framework while providing additional advantages derived from its asynchronous message passing interpretation. In contrast to existing synchronous methods, our model further eliminates the need of stacking multiple message passing layers to reach distant nodes by introducing a suitable neighborhood function. It additionally supports highly efficient training and is able to learn combinations of multi-scale representations.

\section{Neural Message Passing Algorithms}
Neural message passing algorithms follow a \emph{synchronous} neighborhood aggregation scheme. Starting with an initial feature matrix $H^{(0)} \in \RR^{n \times h}$, for $K$ iterations, each node sends a message to each of its neighbors and updates its own state based on the aggregated received messages. Borrowing some notation from \citet{fey2019fast}, we formalize this procedure in Algorithm \ref{alg:smp} where $\mess^{(k)}$ is a \emph{message function}, $\aggr$ is a permutation invariant \emph{aggregation function} and $\comb^{(k)}$ is an \emph{update function}. All of these functions are required to be differentiable. 

\begin{algorithm}[t]
	\caption{Synchronous Message Passing}
	\begin{algorithmic}
		\Require Graph $G$, feature matrix $H^{(0)}$
		\Ensure Aggregated feature matrix $H^{(K)}$
		\For{$k \in [K]$}
		\State \# Send messages
		\For{$i \in V$}
		\For{$j \in \NN_i$}
		\State $\mess_{j \to i}^{(k)} = \mess^{(k)} \left( h_i^{(k - 1)}, h_j^{(k - 1)}, a_{ji} \right)$
		\EndFor
		\EndFor
		\State \# Update node states
		\For{$i \in V$}
		\State $h_i^{(k)} = \comb^{(k)} \left( h_i^{(k - 1)}, \aggr_{j \in \NN_i} \mess_{j \to i}^{(k)} \right)$
		\EndFor
		\EndFor
	\end{algorithmic}
	\label{alg:smp}
\end{algorithm}

One of the most simple and widespread representatives of this framework is the \emph{Graph Convolutional Network (GCN)} \citep{kipf2016semi} which can be defined via

\begin{align}
	\phi^{(k)}_{j \to i} &= \Akipf_{ji} W^{(k)} h_j^{(k)} \\
	\aggr_i^{(k)} &= \sum_{j \in \NN_i} \phi^{(k)}_{j \to i} \\
	\gamma^{(k)}_i &= q^{(k)} \left( \aggr^{(k)}_{i} \right)
\end{align}

such that

\begin{equation}
	H^{(k + 1)} = q^{(k)} \left( \Akipf H^{(k)} W^{(k)} \right)
\end{equation}

where $H^{(0)} = X$, $q^{(k)}$ is a non-linearity (\emph{ReLU} is used for hidden layers, \emph{softmax} for the final prediction layer), $\Akipf = \tilde{D}^{-\nicefrac{1}{2}} \tilde{A} \tilde{D}^{-\nicefrac{1}{2}}$ is a symmetrically normalized adjacency matrix with self-loops and $\tilde{A} = A + I$ with degree matrix $\tilde{D}$. Note that due to self-loops each node also aggregates its own features. Normalization preserves the scale of the feature vectors. In each GCN layer, features are transformed and aggregated from direct neighbors as a weighted sum. 

While various models which can be formulated in this framework have achieved remarkable performance, a general issue with synchronous message passing schemes is that long-range dependencies in the graph are not modeled effectively. If $\NN_i$ denotes the one-hop neighborhood of node $i$ (which is commonly the case), then each message passing iteration expands the receptive field by one hop. For a single node to gather information from another node of distance $K$, $K$ message passing iterations need to be performed for \emph{all} nodes in the graph. Sending a large number of unnecessary messages does not only result in unnecessary computation but further introduces noise to the learned node features. On the same note, \citep{xu2018representation} and \citep{li2018deeper} pointed out an over-smoothing effect. \citet{xu2018representation} showed that with an increasing number of layers, node importance in a GCN converges to the graph's random walk limit distribution, i.e., all local information is lost.

\subsection{Asynchronous Message Passing}
\label{sec:async_mp}

Instead of passing messages along all edges in multiple subsequent rounds, one might consider an \emph{asynchronous} propagation scheme where nodes perform state updates and send messages one after another. 
In particular, pushing natively supports adaptivity, since instead of just pulling information from all neighbors, nodes are able to push and receive important information on demand. This motivates our \emph{push-based message passing framework} (Algorithm \ref{alg:amp}).

\begin{algorithm}[h]
	\caption{Asynchronous Push-based Message Passing}
	\begin{algorithmic}
		\Require Graph $G$, feature matrix $H^{(0)}$
		\Ensure Aggregated feature matrix $H$
		\State Initialize $\Phi_i$ for all $i \in V$
		\While{not converged}
		\State Select next node $i \in V$
		\State \# Update node state
		\State $h_i \leftarrow \comb \left( h_i, \Phi_i \right)$
		\State \# Send messages
		\For{$j \in \NN_i$}
		\State $\mess_{i \to j} = \mess \left( h_i, h_j, \Phi_i, \Phi_j, a_{ij} \right)$
		\State $\Phi_j \leftarrow \aggr \left( \Phi_j, \mess_{i \to j} \right)$
		\EndFor
		\State reset $\Phi_i$
		\EndWhile
	\end{algorithmic}
	\label{alg:amp}
\end{algorithm}

First, it is important to note that each node needs to aggregate incoming messages until it is selected to be updated. For that purpose, we introduce aggregator states $\Phi_i \in \RR^h$ which contain novel unprocessed information for each node. After it is used by a node to update its state and it has pushed messages to its neighbors, the aggregator state is reset until the node receives more information and becomes active again. Further note that the aggregator states naturally lend themselves to serve as a basis for selecting the next node and for a convergence criterion, based on the amount of unprocessed information. The functions $\mess$, $\aggr$ and $\comb$ fulfill the same roles and share the same requirements as their synchronous counterparts. Though not specifically indicated, in principle, different functions may be used for different iterations.

As a particular instance of this framework, we propose \emph{Local Push Message Passing (LPMP)} (Algorithm \ref{alg:appr_mp}). For the next update, the node with the largest aggregator state is selected, since it holds the largest amount of unprocessed information. Similarly, convergence is attained if each node has only a small amount of unprocessed information left. Note that all state updates are additive and no learnable transformations are applied in order to effectively treat long-range dependencies and retain flexibility. Feature transformations may be applied before or after propagation. Further, in each iteration of the outer loop, only the features of node $k$ are diffused through the graph in order to avoid excessive smoothing which might occur when multiple features are propagated at the same time over longer distances in the graph. Also, all iterations of the outer loop are independent of each other and can be performed in parallel. Remaining details of the algorithm will be motivated and explained below.

We further wish to point out that the synchronous framework does not consider any notion of convergence but instead introduces a hyper-parameter for the number of message passing iterations. An early work on \emph{Graph Neural Network (GNN)} \citep{scarselli2009graph} applies contraction mappings and performs message passing until node states converge to a fixed point. However, neighborhood aggregation is still performed synchronously.

Finally, further instances of the general push-based message passing framework may be considered in future work. We focus on this particular algorithm due to its nice interpretation in terms of existing push algorithms (as detailed below), its favorable properties and since we observed it to perform well in practice.

\begin{algorithm}[t]
	\caption{Local Push Message Passing (LPMP)}
	\begin{algorithmic}
		\Require Graph $G$, feature matrix $H^{(0)}$, parameters $\alpha \in (0, 1)$, $\varepsilon > 0$
		\Ensure Aggregated feature matrix $H$
		\State Initialize dense matrix $H = zeros(n, h)$
		\For{$k \in V$}
		\State $\Phi_i = \delta_{i,k} h_i$ for all $i \in V$
		\While{$ \max_{i \in V} ||\Phi_i|| > \varepsilon ||h_k||$}
		\State $h_i \leftarrow h_i + \alpha \Phi_i$
		\For{$j \in \NN_i$}
		\State $\Phi_j \leftarrow \Phi_j + \frac{1 - \alpha}{d_j} \Phi_i$
		\EndFor
		\State $\Phi_i = 0$
		\EndWhile
		\EndFor
	\end{algorithmic}
	\label{alg:appr_mp}
\end{algorithm}

\label{sec:message_passing}

\section{Pushing Networks}
The LPMP algorithm described above is inspired by local \emph{push} algorithms for computation of \emph{Approximate Personalized PageRank (APPR)} \citep{jeh2003scaling,berkhin2006bookmark} and, in particular, we will show in the following how it can be equivalently described as a single synchronous message passing iteration using sparse APPR neighborhoods. Thus, the proposed message passing scheme effectively combines the advantages of existing synchronous algorithms with the benefits of asynchronous message passing described above.

\subsection{Personalized Node Neighborhoods}

\emph{Personalized PageRank (PPR)} refers to a localized variant of \emph{PageRank} \citep{page1999pagerank} where random walks are restarted only from a certain set of nodes. We consider the special case in which the starting distribution is a unit vector, i.e., when computing PPR-vector $\pi_i$ of node $i$, walks are always restarted at $i$ itself. Formally, $\pi_i$ can be defined as the solution of the linear system

\begin{equation}
	\pi_i = \alpha e_i + (1 - \alpha) \pi_i A_{rw}
	\label{eq:ppr}
\end{equation}
 
where $e_i \in \RR^n$ denotes the $i$th unit vector, $A_{rw} = D^{-1} A$ denotes the random walk transition matrix of $G$, and the restart probability $\alpha \in (0, 1)$ controls the locality, where a larger value leads to stronger localization. The PPR vectors for all nodes can be stored as rows of a PPR-matrix $\Pi \in \RR^{n \times n}$. Intuitively, $\pi_{ij}$ corresponds to the probability that a random walk starting at $i$ stops at $j$ where the expected length of the walk is controlled by $\alpha$. The vector $\pi_i$ can be interpreted as an importance measure for node $i$ over all other nodes where $\pi_{ij}$ measures the importance of $j$ for $i$. Since these measures are not sparse and global computation of $\Pi$ would require $\OO(n^2)$ operations, we consider local computation of APPR instead. In particular, we refer to the \emph{Reverse Local Push} algorithm \citep{andersen2007local}, since it comes with several useful theoretical properties. Complexity of computing the whole matrix $P$ is reduced to $\OO(\nicefrac{n}{\alpha \varepsilon})$ \citep{andersen2007local}, i.e., linear in the number of nodes. The parameter $\varepsilon > 0$ controls the quality of approximation, sparsification and runtime where a larger value leads to sparser solutions.
For a more in-depth discussion, we refer to \cite{andersen2007local}.

\subsection{PushNet}

Based on the above neighborhood function, we propose the following neural message passing algorithm:

\begin{definition}[PushNet]
	Let $f$ and $g$ be MLPs parametrized by $\theta_f$ and $\theta_g$, respectively, $h_1, h_2, h_3$ denote hidden dimensions, $P = \left[ P^{(\alpha_1)}, \dots, P^{(\alpha_K)} \right] \in \RR^{K \times n \times n}$ be a tensor storing precomputed APPR matrices for different scales $\alpha_1 \geq \dots \geq \alpha_K$ and $\saggr$ denote a differentiable \emph{scale aggregation function}. Given input features $X \in \RR^{n \times d}$, the layers of \emph{PushNet} are defined as
	
	\begin{align}
	H^{(0)} &= f\left(X; \theta_f\right) & &\in \RR^{n \times h_1} \\
	H^{(1)} &= P H^{(0)} & &\in \RR^{K \times n \times h_1} \\
	H^{(2)} &= \saggr \left( H^{(1)} \right) & &\in \RR^{n \times h_2} \\
	H^{(3)} &= g\left(H^{(2)}; \theta_g\right) & &\in \RR^{n \times h_3}
	\end{align}
	
\end{definition}

In most cases, $h_3=c$, such that $H^{(3)}$ provides the final predictions for each node over $c$ classes. In general, PushNet might also be applied to different graph learning problems such as graph classification, where learned node representations are pooled and labels are predicted for whole graphs. However, we leave these further applications to future work. 

To draw the connection between synchronous and asynchronous message passing, we show that the base variant of PushNet with no feature transformations and a single scale $\alpha$ is equivalent to LPMP (Algorithm \ref{alg:appr_mp}):

\begin{theorem}
	Let $\alpha \in (0,1)$ and $\varepsilon > 0$ be fixed, $f,g,\saggr$ be identity functions and $K=1$. Then $H^{(3)} = H$	where $H^{(3)} = PX$ is the output of PushNet and $H$ is the output of LPMP.
	\label{thm:equi}
\end{theorem}

The main idea is that instead of propagating features directly as in LPMP, we can first propagate scalar importance weights as in Reverse Local Push and then propagate features in a seconds step. Thus, all discussion on LPMP are directly applicable to PushNet, including adaptivity, effective treatment of long-range dependencies and avoidance of over-smoothing. We wish to point out that an additional interpretation of adaptivity can be derived from the perspective of PushNet: APPR-induced neighborhoods of different nodes are sparse and directly exclude irrelevant nodes from consideration, in contrast to commonly used $k$-hop neighborhoods. In this sense, APPR is adaptive to the particular source node. To the best of our knowledge, no existing neural message passing algorithm shares this property.

In practice, it is favorable to not propagate features using LPMP, but to pre-compute APPR matrices such that features are propagated only once along all non-zero APPR entries and there is no need to propagate gradients back over long paths of messages. Thus, PushNet benefits from the existing synchronous neural message passing framework while providing additional advantages derived from its asynchronous interpretation.

\subsection{Learning Multi-Scale Representations}

Additional properties of PushNet compared to LPMP include feature transformations $f$ and $g$ which may be applied before and after feature propagation. Since the optimal neighborhood size cannot be assumed to be the same for each node and patterns might be observed at multiple scales, we additionally propagate features over different localities by varying the restart probability $\alpha$. The multi-scale representations are then aggregated per node into a single vector such that the model learns to combine different scales for a given node. In particular, we consider the following scale aggregation functions:

\begin{itemize}
	\item \textbf{sum}: Summation of multi-scale representations. Intuitively, sum-aggregation corresponds to an unnormalized average with uniform weights attached to all scales.
	\begin{equation}
		\saggr \left( H^{(1)} \right) = \sum_{k \in [K]} P^{(\alpha_k)} H^{(0)} \in \RR^{n \times h_1}
	\end{equation} 
	Note that due to distributivity, PushNet with sum aggregation reduces to propagation with a single matrix $P = \sum_{k \in [K]} P^{(\alpha_k)}$, i.e., features can be propagated and additively combined over an arbitrary number of different scales at the cost of only a single propagation. Thereby, the non-zero entries in $P$ are given by $\nz{P} = \bigcup_{k \in [K]} \nz{P^{(\alpha_k)}}$. However, usually the number of non-zero entries $\nnz{P}$ will be close to $\nnz{P^{(\alpha_K)}}$, since nodes considered at a smaller scale will most often also be considered at a larger scale. Thus, complexity will be dominated by the largest scale considered.
	\item \textbf{max}: Element-wise maximum of multi-scale representations. The most informative scale is selected for each feature individually. This way, different features may correspond to more local or more global properties.
	\begin{equation}
	\saggr \left( H^{(1)} \right) = \max_{k \in [K]} P^{(\alpha_k)} H^{(0)} \in \RR^{n \times h_1}
	\end{equation}
	\item \textbf{cat}: Concatenation of multi-scale representations. Scale combination is learned in subsequent layers. The implied objective is to learn a scale aggregation function which is globally optimal for all nodes.
	\begin{equation}
	\saggr \left( H^{(1)} \right) = \concat_{k \in [K]} P^{(\alpha_k)} H^{(0)} \in \RR^{n \times (K \cdot h_1)}
	\end{equation}
\end{itemize}

\subsection{The PushNet Model Family}

We wish to point out several interesting special cases of our model.
In our default setting, prediction layers will always be dense with a \emph{softmax} activation. If hidden layers are used, we use a single dense layer with \emph{ReLU} activation.

\begin{itemize}
	\item \textbf{PushNet.} The general case in which $f$ and $g$ are generic MLPs. As per default, $f$ is a single dense hidden layer and $g$ is a dense prediction layer.
	\item $\mathbf{f = id.}$ No feature transformation is performed prior to propagation. In this case, $H^{(2)}$ needs to be computed only once and can then be cached, making learning extremely efficient. The following sub-cases are of particular interest:
	\begin{itemize}
		\item \textbf{PushNet-PTP.} The sequence of operations is "push -- transform -- predict". In this case, $g$ is a generic MLP, consisting of 2 layers per default.
		\item \textbf{PushNet-PP.} The model performs operations "push -- predict" and uses no hidden layers. Predictions can be interpreted as the result of logistic regression on aggregated features. This version is similar to SGC \citep{wu2019simplifying} with the difference that SGC does not consider multiple scales and propagates over $k$-hop neighborhoods.
		\item \textbf{LPMP.} In this setting, $g = id$ and $K = 1$, i.e., no feature transformations are performed and features are aggregated over a single scale. This setting corresponds to LPMP, cf. Theorem \ref{thm:equi}. Note that this model describes only feature propagation, no actual predictions are made.
	\end{itemize}
	\item \textbf{PushNet-TPP.} The setting is $h_1 = c$ and $g = id$, such that the model first predicts class labels for each node and then propagates the predicted class labels. This setting is similar to APPNP \citep{klicpera2018predict} but with some important differences. APPNP considers only a single fixed scale and does not propagate over APPR neighborhoods. Instead, multiple message passing layers are stacked to perform a power iteration approximation of PPR. The resulting receptive field is restricted to $k$-hop neighborhoods. Note that \emph{cat} aggregation is not applicable here.
\end{itemize}

\vspace{-2em}
\subsection{Comparison with Existing Neural Message Passing Algorithms}

Existing message passing algorithms have explored different concepts of node importance, i.e., weights used in neighborhood aggregation. While GCN \citep{kipf2016semi} and other GCN-like models use normalized adjacency matrix entries in each layer, \emph{Simplified Graph Convolution (SGC)} \citep{wu2019simplifying} aggregates nodes over $k$-hop neighborhoods in a single iteration using a $k$-step random walk matrix. \emph{Approximate Personalized Propagation of Neural Predictions (APPNP)} \citep{klicpera2018predict} also relies on a $k$-step random walk matrix but with restarts which can be equivalently interpreted as a power iteration approximation of the PPR matrix. \emph{Graph Attention Network (GAT)} \citep{velivckovic2017graph} learns a similarity function which computes a pairwise importance score given two nodes' feature vectors. All of the above methods aggregate features over fixed $k$-hop neighborhoods. PushNet on the other hand aggregates over sparse APPR neighborhoods using the corresponding importance scores.

\begin{table}
	\centering
\resizebox{\columnwidth}{!}{%
	\begin{tabular}{lrrrrrr}
	    \toprule
		& $\mathbf{|V|}$ & $\mathbf{|E|}$ & $\mathbf{d}$ & $\mathbf{c}$ & $\mathbf{avgSP}$ & $\mathbf{maxSP}$ \\
		\toprule
		\textbf{CiteSeer} & 2120 & 3679 & 3703 & 6 & 9.33 & 28 \\
		\textbf{Cora} & 2485 & 5069 & 1433 & 7 & 6.31 & 19 \\
		\textbf{PubMed} & 19717 & 44324 & 500 & 3 & 6.34 & 18 \\
		\textbf{Coauthor CS} & 18333 & 81894 & 6805 & 15 & 5.43 & 24 \\
		\textbf{Coauthor Physics} & 34493 & 247962 & 8415 & 5 & 5.16 & 17 \\
        \bottomrule
	\end{tabular}%
}
\caption{Dataset statistics including average and maximum shortest path lengths considering only the largest connected component of each graph.}
\label{table:datasets}
\end{table}

Multi-scale representations have been considered in \emph{Jumping Knowledge Networks (JK)} \citep{xu2018representation} where intermediate representations of a GCN or GAT base network are combined before prediction. The original intention was to avoid over-smoothing by introducing these skip connections. Similarly, \citep{liao2019lanczosnet} concatenate propagated features at multiple selected scales. LD \citep{faerman2018semi} uses APPR to compute local class label distributions at multiple scales and proposes different combinations but is limited to unattributed graphs. PushNet varies the restart probability in APPR to compute multi-scale representations and combines them using simple and very efficient aggregation functions. APPR-Roles \citet{borutta2019structural} also employs APPR but to compute structural node embeddings in an unsupervised setting.
The idea of performing no learnable feature transformations prior to propagation was explored already in SGC \cite{wu2019simplifying}. It allows for caching propagated features, resulting in very efficient training, and is also used by PushNet-PTP and PushNet-PP.
LD \citep{faerman2018semi} explored the idea of propagating class labels instead of latent representations. Similarly, APPNP \citep{klicpera2018predict} and PushNet-TPP propagate predicted class labels.
In parallel to our research, \citep{bojchevski2019scaling} have developed PPRGo, a model similar to Pushnet-TPP, but the authors focus on scalability aspects rather than asynchronous message passing.

To the best of our knowledge, the only existing work considering asynchronous neural message passing is SSE \citep{dai2018learning}. Compared to PushNet, SSE is pull-based, i.e., in each iteration a node pulls features from all neighbors and updates its state, until convergence to steady node states. To make learning feasible, stochastic training is necessary. Further, the work focuses on learning graph algorithms for different tasks and results for semi-supervised node classification were not very competitive. In contrast, PushNet offers very fast deterministic training and adaptive state updates due to a push-based approach.

\vspace{-0.75em}
\section{Experiments}
\vspace{-0.5em}
We compare PushNet and its variants against six state-of-the-art models, GCN \citep{kipf2016semi}, GAT \citep{velivckovic2017graph}, JK \citep{xu2018representation} with base model GCN and GAT, SGC \citep{wu2019simplifying}, \emph{Graph Isomorphism Network (GIN)} \citep{xu2018powerful} and APPNP \citep{klicpera2018predict} on five established node classification benchmark datasets. For better comparability, all models were implemented using \emph{PyTorch Geometric} \citep{fey2019fast} and trained on a single NVIDIA GeForce GTX 1080 Ti GPU. We make an implementation of our models publicly available. \footnote{\url{https://github.com/buschju/pushnet}}

\begin{table}
\resizebox{\columnwidth}{!}{%
	\centering
	\begin{tabular}{lrrrr}
		\toprule
		 & \textbf{Hidden size} & \textbf{Learning rate} & \textbf{Dropout} & \textbf{$L_2$ reg. strength} \\
		\midrule
		\textbf{GCN} & 64 & 0.01 & 0.5 & 0.001 \\
		\textbf{GAT} & 64 & 0.01 & 0.6 & 0.001 \\
		\textbf{SGC} & --- & 0.01 & --- & 0.0001 \\
		\textbf{GIN} & 64 & 0.01 & 0.6 & --- \\
		\textbf{JK-GCN} (cat) & 64 & 0.01 & 0.4 & 0.001 \\
		\textbf{JK-GAT} (cat) & 64 & 0.01 & 0.4 & 0.001 \\
		\textbf{APPNP} ($\alpha=0.1, K=10$) & 64 & 0.01 & 0.5 & 0.01 \\
		\midrule
		\textbf{PushNet} (sum) & 64 & 0.005 & 0.5 & 0.01 \\
		\textbf{PushNet-PTP} (sum) & 64 & 0.005 & 0.3 & 0.1 \\
		\textbf{PushNet-PP} (sum) & --- & 0.01 & 0.6 & 0.001 \\
		\textbf{PushNet-TPP} (sum) & 32 & 0.01 & 0.5 & 0.01 \\
		\bottomrule
	\end{tabular}
	}
\caption{Optimal hyper-parameters for all models as determined by a grid search on CiteSeer and Cora. Values in parentheses indicate optimal model-specific hyper-parameters.}
\label{tab:optimal_parameters}
\end{table}

\vspace{-0.75em}
\subsection{Datasets}
\vspace{-0.75em}

Experiments are performed on semi-supervised text classification benchmarks. In particular, we consider three citation networks, \emph{CiteSeer} and \emph{Cora} from \citep{sen2008collective} and \emph{PubMed} from \citep{namata2012query}, and two co-authorship networks, \emph{Coauthor CS} and \emph{Coauthor Physics} from \citep{shchur2018pitfalls}.
Statistics of these datasets are summarized in Table \ref{table:datasets}.

\vspace{-0.75em}
\subsection{Experimental Setup}
\vspace{-0.5em}

For the sake of an unbiased and fair comparison, we follow a rigorous evaluation protocol, similarly as in \citep{shchur2018pitfalls} and \citep{klicpera2018predict}. \footnote{Note that results for competing methods might differ from those reported in related work due to a different experimental setup.} We restrict all graphs to their largest connected components and $L1$-normalize all feature vectors. Self-loops are added and different normalizations are applied to the adjacency matrices individually for each method as proposed by the respective authors. For each dataset, we sample 20 nodes per class for training and 500 nodes for validation. The remaining nodes are used as test data. Each model is evaluated on 20 random data splits with 5 random initializations, resulting in 100 runs per model and dataset. Using the same random seed for all models ensures that all models are evaluated on the same splits.

Model architectures including sequences and types of layers, activation functions, locations of dropout and $L2$-regularization are fixed as recommended by the respective authors. All remaining hyperparameters are optimized per model by selecting the parameter combination with best average accuracy on CiteSeer and Cora validation sets. Final results are reported only for the test sets using optimal parameters. 
All models are trained with \emph{Adam} \citep{kingma2014adam} using default parameters and early stopping based on validation accuracy and loss as in \citep{velivckovic2017graph} with a patience of 100 for a maximum of 10000 epochs.

For all PushNet variants, we fix the architecture as described in the previous section. Dropout is applied to all APPR matrices and to the inputs of all dense layers. However, for PushNet-PTP and PushNet-PP dropout is only applied after propagation, such that propagated features can be cached. $L2$-regularization is applied to all dense layers. As a default setting, we consider three different scales $\alpha \in \{ 0.2, 0.1, 0.05 \}$ and $\varepsilon = 1e-5$. Due to memory constraints, we use $\varepsilon = 1e-4$ on Physics dataset for all PushNet variants and on CS and PubMed datasets for PushNet and PushNet-TPP. We further add self-loops to the adjacency matrices and perform symmetric normalization as in GCN. All APPR-matrices are $L1$-normalized per row.
We use the following parameter grid for tuning hyper-parameters of all models:

\vspace{-2pt}
\begin{itemize}
	\item Number of hidden dimensions: [8, 16, 32, 64]
	\item Learning rate: [0.001, 0.005, 0.01]
	\item Dropout probability: [0.3, 0.4, 0.5, 0.6]
	\item Strength of L2-regularization: [1e-4, 1e-3, 1e-2, 1e-1]
\end{itemize}
\vspace{-2pt}

Except for APPNP, all competitors use $K = 2$ propagation layers. JK and GIN use an additional dense layer for prediction. For GAT layers, the number of attention heads is fixed to 8. Optimal hyper-parameters for all models are reported in Table \ref{tab:optimal_parameters}.

\begin{figure*}[!ht]
	\begin{subfigure}[b]{0.33\textwidth}
		\centering
		\includegraphics[width=\textwidth]{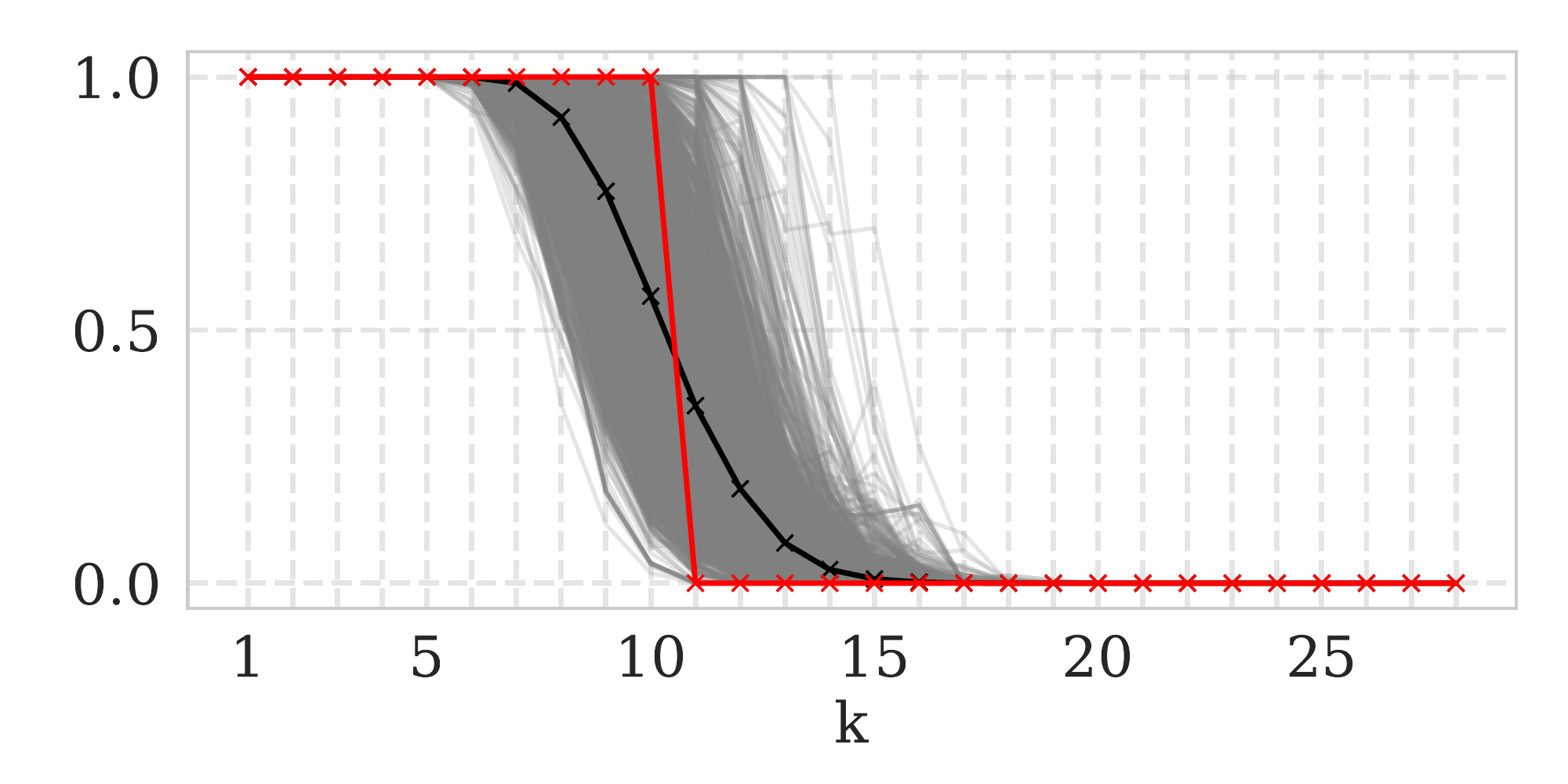}
		\caption{CiteSeer ($\alpha=0.05$, $\varepsilon=1e-5$)}
		\label{fig:appr_size}
	\end{subfigure}
	\begin{subfigure}[b]{0.33\textwidth}
		\centering
		\includegraphics[width=\textwidth]{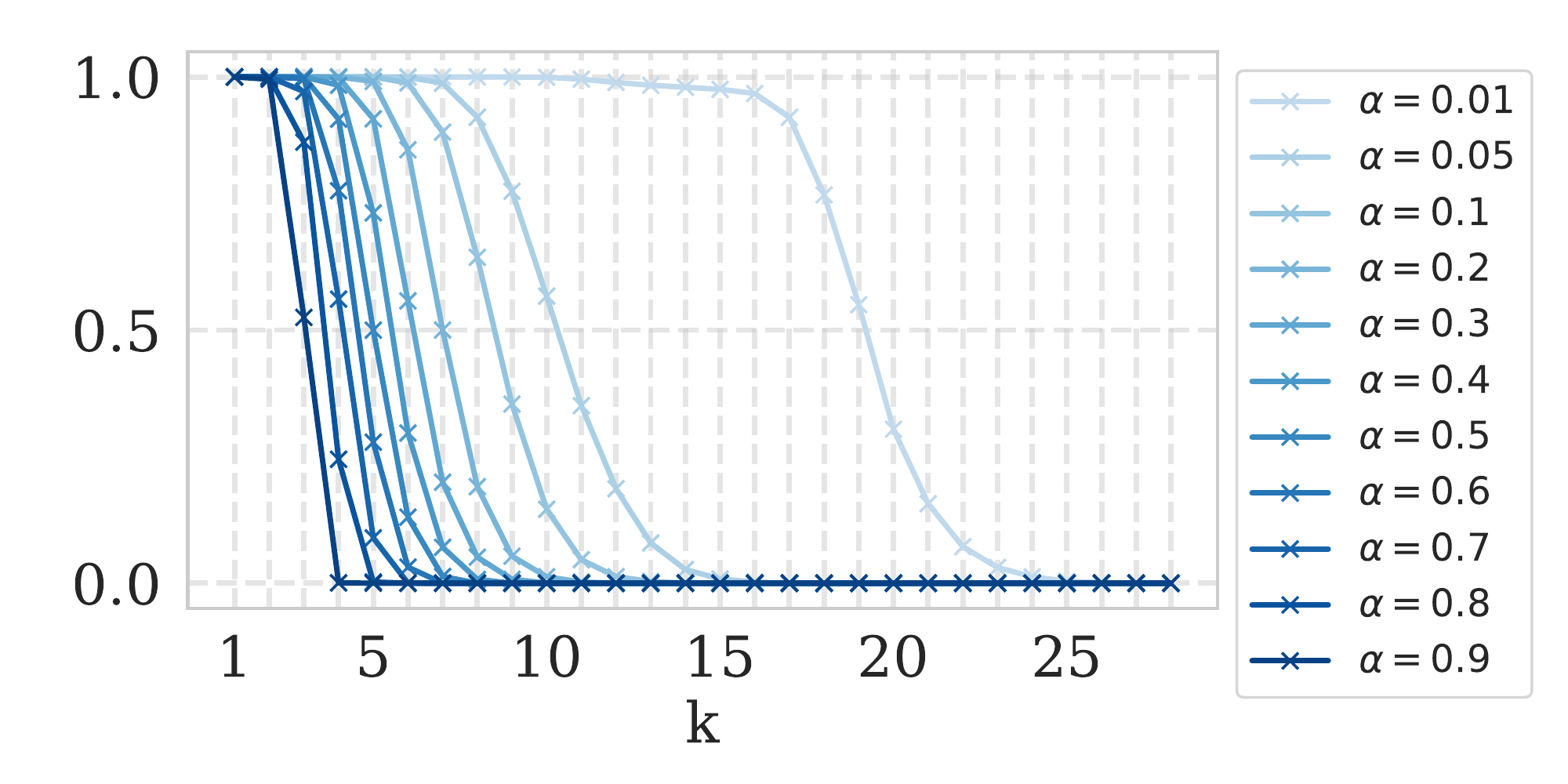}
		\caption{CiteSeer ($\varepsilon=1e-5$)}
		\label{fig:appr_size_alphas}
	\end{subfigure}
	\begin{subfigure}[b]{0.33\textwidth}
		\centering
		\includegraphics[width=\textwidth]{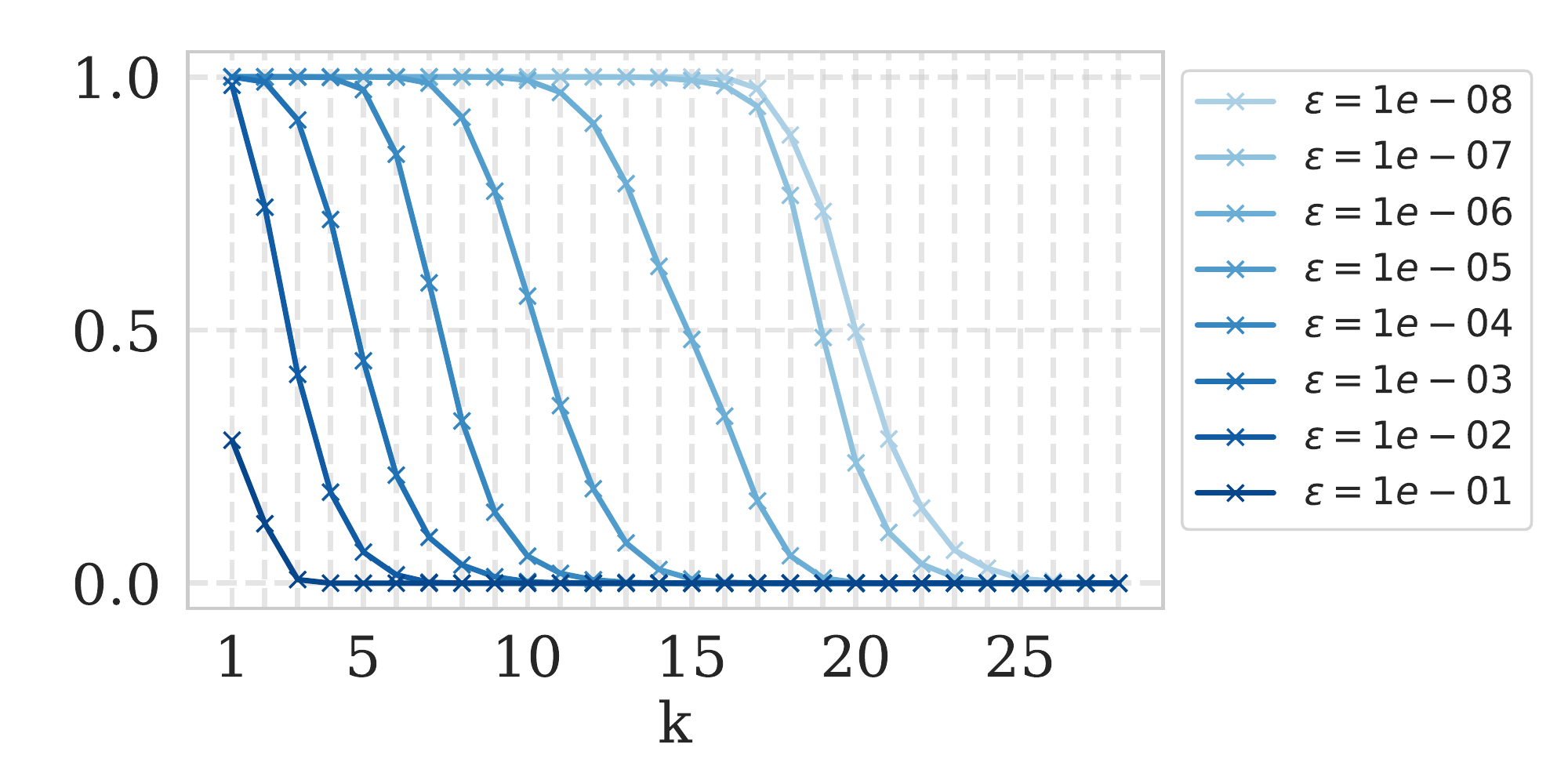}
		\caption{CiteSeer ($\alpha=0.05$)}
		\label{fig:appr_size_epsilons}
	\end{subfigure}
	\caption{Fraction of $k$-hop neighbors contained in APPR-neighborhoods for CiteSeer. (\ref{fig:appr_size}) shows all APPR neighborhoods for a fixed setting, including mean and the closest $k$-neighborhood for reference. (\ref{fig:appr_size_alphas}) and (\ref{fig:appr_size_epsilons}) demonstrate localization depending on $\alpha$ and sparsification depending on $\varepsilon$, respectively.}
	\label{fig:appr_neighborhoods}
\end{figure*}

\begin{table*}[t]
	\centering
	\resizebox{\textwidth}{!}{%
		\begin{tabular}{llllllll|llll}
\toprule
{} &      \textbf{GCN} &      \textbf{GAT} &   \textbf{JK-GCN} &   \textbf{JK-GAT} &      \textbf{SGC} &      \textbf{GIN} &    \textbf{APPNP} &  \textbf{PushNet} &       \textbf{PushNet-PTP} & \textbf{PushNet-PP} &       \textbf{PushNet-TPP} \\
\midrule
\textbf{CiteSeer        } &  $72.82 \pm 1.48$ &  $73.82 \pm 1.35$ &  $71.09 \pm 1.66$ &  $71.76 \pm 1.27$ &  $73.91 \pm 1.30$ &  $70.81 \pm 1.61$ &  $74.36 \pm 1.44$ &  $75.08 \pm 0.99$ &  $\mathbf{75.19 \pm 1.15}$ &    $75.17 \pm 1.32$ &           $75.01 \pm 1.11$ \\
\textbf{Cora            } &  $81.07 \pm 1.43$ &  $82.12 \pm 1.41$ &  $79.57 \pm 1.63$ &  $80.10 \pm 1.52$ &  $80.13 \pm 2.15$ &  $80.24 \pm 1.54$ &  $83.58 \pm 1.03$ &  $84.12 \pm 1.08$ &           $83.41 \pm 1.24$ &    $81.52 \pm 1.40$ &  $\mathbf{84.23 \pm 1.26}$ \\
\textbf{PubMed          } &  $78.29 \pm 1.48$ &  $78.21 \pm 1.60$ &  $77.23 \pm 2.01$ &  $77.59 \pm 2.25$ &  $77.00 \pm 1.78$ &  $77.19 \pm 1.75$ &  $79.61 \pm 2.98$ &  $79.80 \pm 1.39$ &  $\mathbf{80.22 \pm 1.27}$ &    $77.52 \pm 2.05$ &           $80.10 \pm 1.33$ \\
\textbf{Coauthor CS     } &  $91.64 \pm 0.62$ &  $90.20 \pm 0.75$ &  $91.60 \pm 0.54$ &  $92.20 \pm 0.43$ &  $91.27 \pm 0.58$ &  $91.46 \pm 0.54$ &  $91.10 \pm 1.12$ &  $92.40 \pm 0.52$ &           $92.37 \pm 0.40$ &    $91.04 \pm 0.76$ &  $\mathbf{92.54 \pm 0.34}$ \\
\textbf{Coauthor Physics} &  $93.42 \pm 0.63$ &  $93.43 \pm 0.50$ &  $93.49 \pm 0.56$ &     $o.o.m.$ &     $o.o.m.$ &  $93.79 \pm 0.49$ &  $93.96 \pm 0.45$ &  $94.01 \pm 0.53$ &           $93.97 \pm 0.48$ &    $93.67 \pm 0.55$ &  $\mathbf{94.09 \pm 0.47}$ \\
\bottomrule
\end{tabular}

	}
	\caption{Accuracy/micro-F1 scores on semi-supervised node classification datasets in terms of mean and standard deviation over 100 independent runs. JK-GAT and SGC are out of GPU memory on the largest dataset, Coauthor Physics.}
	\label{tab:results_acc}
\end{table*}

\begin{table}
	\centering
	\resizebox{\columnwidth}{!}{%
		\begin{tabular}{llllll}
\toprule
{} & \textbf{CiteSeer} & \textbf{Cora} & \textbf{PubMed} & \textbf{Coauthor CS} & \textbf{Coauthor Physics} \\
\midrule
\textbf{Accuracy} &          4.18e-08 &      3.37e-07 &        1.20e-02 &             4.62e-11 &                  3.45e-03 \\
\textbf{Macro-F1} &          3.91e-10 &      2.79e-04 &        2.22e-02 &             9.22e-12 &                  2.12e-07 \\
\bottomrule
\end{tabular}

	}
	\caption{P-values according to a Wilcoxon signed-rank test comparing the best of our models with the best competitor on all datasets with respect to accuracy/micro-F1 and macro-F1.}
	\label{tab:p_values}
\end{table}

\subsection{Node Classification Accuracy}

\begin{figure*}[t]
	\includegraphics[width=\textwidth]{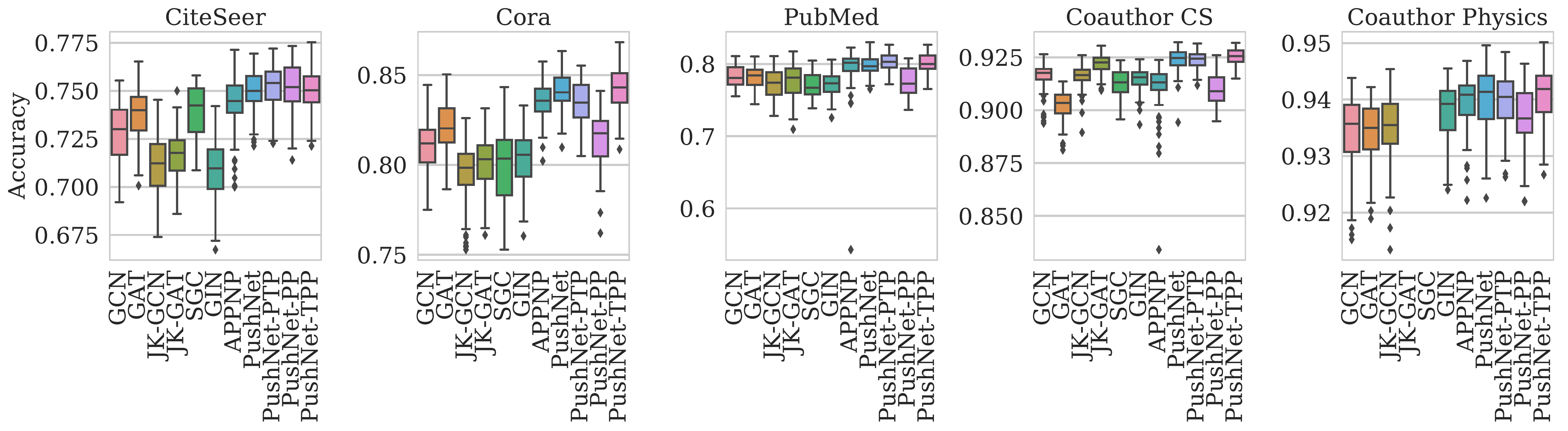}
	\caption{Accuracy/micro-F1 scores on semi-supervised node classification datasets aggregated over 100 independent runs.}
	\label{fig:boxplots_acc}
\end{figure*}

Accuracy/micro-F1 scores for all datasets are provided in Table \ref{tab:results_acc}. It can be observed that our models consistently provide best results on all datasets and that the strongest model, PushNet-TPP, outperforms all competitors on all datasets. 
Improvements of our best model compared to the best competing model are statistically significant with $P < .05$ on all datasets according to a Wilcoxon signed-rank test. \footnote{In fact, results are significant with $P < 0.01$ on all datasets except PubMed.}
P-values for all datasets are reported in Table \ref{tab:p_values}.
On CiteSeer and PubMed, PushNet-PTP is able to push performance even further. PushNet with feature transformations before and after propagation is less performant but still outperforms all competitors on all datasets. PushNet-PP, the most simple of our models, performs worst as expected. However, it is still competitive, outperforming all competitors on CiteSeer. Boxplots shown in Figure \ref{fig:boxplots_acc} indicate that our models generally exhibit small variance and are less prone to produce outlier scores.

We argue that improvements over existing methods are primarily due to push-based propagation. Figure \ref{fig:appr_size} compares APPR neighborhoods with $k$-hop neighborhoods in terms of the fraction of $k$-neighbors considered. It can be seen that $k$-neighborhoods used by competitors draw a sharp artificial boundary while APPR adaptively selects nodes from larger neighborhoods and discards nodes from smaller ones, \emph{individually for each source node}. Visually, deviations left to the boundary correspond to discarded irrelevant nodes, while deviations on the right hand side indicate additional nodes beyond the receptive field of competitors that can be leveraged by our method. Stacking more message passing layers to reach these nodes would degrade performance due to overfitting as demonstrated in \citep{xu2018representation} and \citep{li2018deeper}.

Among the competing methods, APPNP performs best in general, providing best baseline performance on all datasets but CS where JK-GAT achieves best results. GAT outperforms GCN on three datasets, CiteSeer, Cora and Physics. JK performs worse than its respective basemodel in most cases. Similar observations were already made in \citep{klicpera2018predict}. SGC mostly performs worse than GCN due to its simplicity, outperforming it only on CiteSeer. GIN also provides worse results than GCN in most cases, possibly due to overfitting caused by larger model complexity. It outperforms GCN only on Physics, providing results similar to APPNP.

On CiteSeer, models using cached features perform very well, even the simple models SGC and PushNet-PP which effectively perform linear regression on propagated raw features provide superior performance. On the remaining datasets, the additional feature transformation provided by PushNet-PTP is necessary to guarantee high accuracy. 

Macro-F1 scores reveal similar insights and are ommitted due to space constraints.
\vspace{-8pt}

\begin{figure*}[t]
	\includegraphics[width=\textwidth]{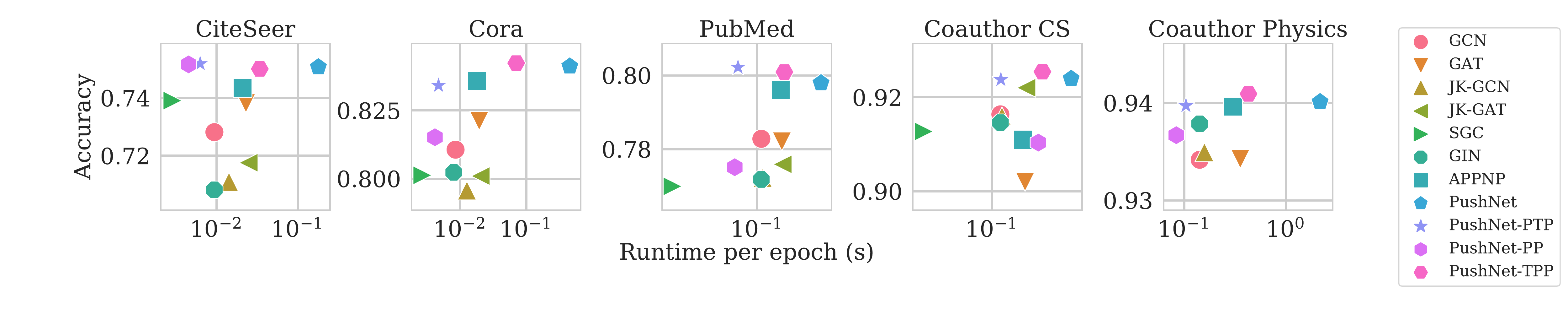}
	\caption{Comparison w.r.t. average runtime per epoch in seconds vs. average accuracy.}
	\label{fig:runtime_acc}
\end{figure*}

\subsection{Runtime}

Figure \ref{fig:runtime_acc} compares all methods based on average runtime per epoch and accuracy. \footnote{We note that for APPNP and all PushNet variants, (A)PPR matrix computation is not included in runtime analysis such that runtime comparison is solely based on propagation, transformation and prediction for all compared models. We consider (A)PPR computation as a preprocessing step, since it is only required to be performed once per graph and can then be reused by all models for this graph. Computation is very fast for each of the considered graphs and can be performed on CPU.} SGC has lowest runtime on all datasets but runs out of memory on Physics since it propagates raw features over $k$-hop neighborhoods. PushNet-PP performs second fastest, followed by PushNet-PTP which generally provides a good tradeoff between runtime and accuracy. PushNet and PushNet-TPP are slower than competitors but still provide comparable runtime at a superior level of accuracy. Among the competitors, APPNP, GAT and JK-GAT require most computation time. JK-GAT also runs out of memory on Physics. 

\begin{figure}[t]
	\centering
	\includegraphics[width=\columnwidth]{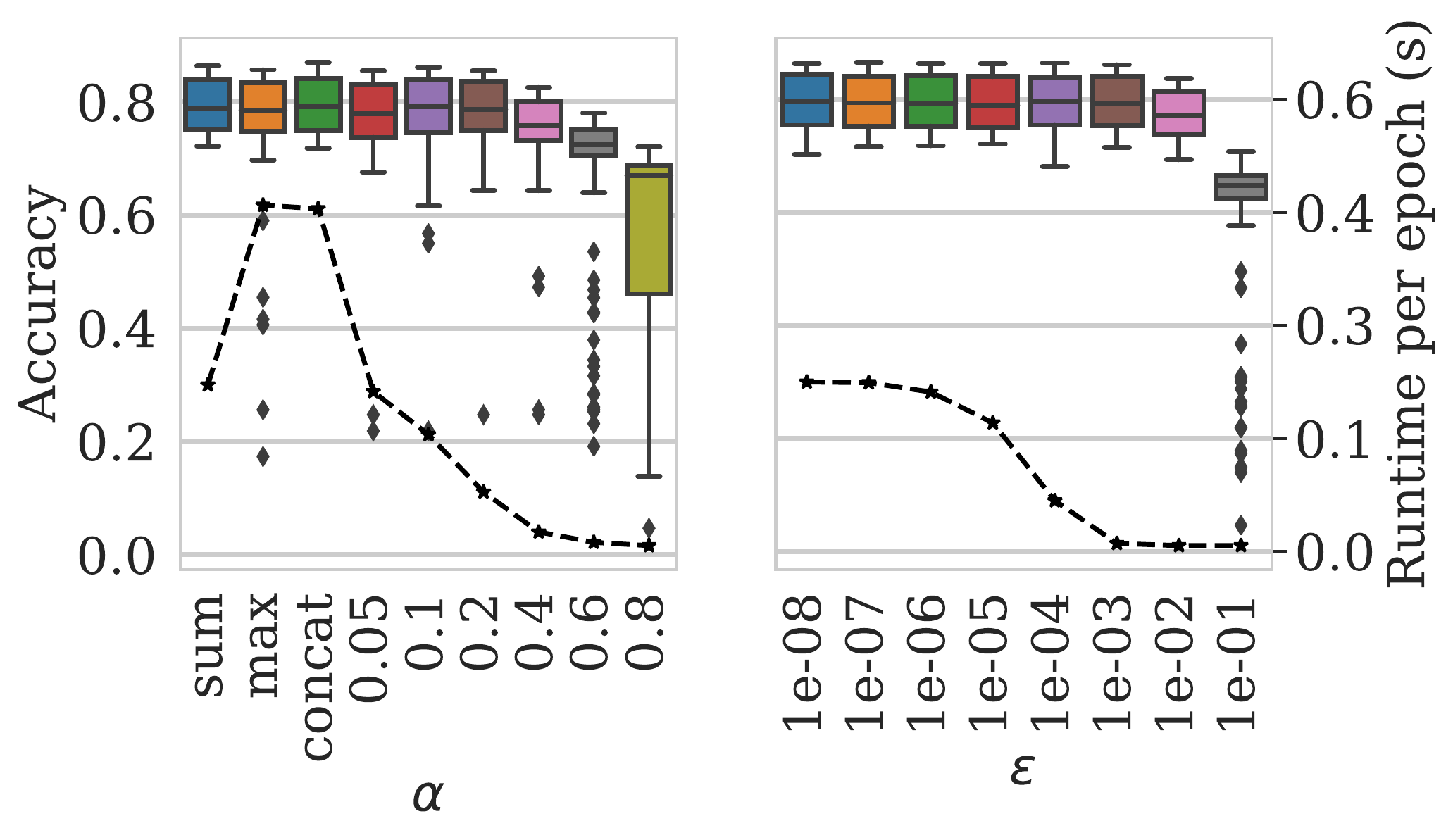}
	\caption{Effect of varying $\alpha$ for fixed $\varepsilon = 1e-5$ (left) and varying $\varepsilon$ for fixed $\alpha \in \{ 0.05, 0.1, 0.2 \}$ with \emph{sum} aggregation (right). Dashed lines indicate average runtime per epoch in seconds.
}
	\label{fig:vary_alpha_epsilon}
\end{figure}

\subsection{Influence of Locality}

To study the influence of the locality parameter $\alpha$ on the performance of our models, we run experiments with our base model PushNet with various single $\alpha$ values and different aggregations of multiple values. Figure \ref{fig:appr_size_alphas} illustrates how the fraction of $k$-neighbors considered for propagation varies with $\alpha$. Generally, a larger value leads to stronger localization and the shape gets closer to a step function as for $k$-neighborhoods. Figure \ref{fig:vary_alpha_epsilon} additionally shows the average performance on CiteSeer and Cora. 
For single $\alpha$, small values in $\{ 0.05, 0.1, 0.2 \}$ achieve best accuracy. For larger $\alpha$, runtime drops considerably but at the cost of decreased accuracy and larger variance. Among multi-scale aggregations, \emph{sum} performs best. It slightly improves the performance over single alphas and provides additional robustness, producing smaller variance and no outlier scores. Runtime is very close to the smallest single $\alpha$ considered. The remaining aggregation functions lead to increased runtime and \emph{max} does not even lead to an increase of accuracy.

\subsection{Influence of Sparsity}

Similarly as $\alpha$, the approximation threshold $\varepsilon$ controls the effective neighborhood size considered for propagation. We study the effect on our models with a similar setup as above. Figure \ref{fig:appr_size_epsilons} illustrates how a larger value of $\varepsilon$ leads to stronger sparsification of APPR neighborhoods. Variation leads to a shift of the curve, indicating that neighbors with small visiting probabilities are discarded mostly from $k$-neighborhoods with moderate or large $k$. Figure \ref{fig:vary_alpha_epsilon} shows that accuracy remains relatively stable on CiteSeer and Cora, decreasing monotonically for increasing $\varepsilon$. Simultaneously, runtime decreases steadily. 
While smaller $\varepsilon$ provide marginally better accuracy, our results suggest that $\varepsilon$ may be increased safely to allow for faster runtime and to account for limited GPU memory.

\section{Conclusion}
We presented a novel push-based asynchronous neural message passing algorithm which allows for efficient feature aggregation over adaptive node neighborhoods. A multi-scale approach additionally leverages correlations on increasing levels of locality and variants of our model capture different inductive bias. Semi-supervised node classification experiments on five real-world benchmark datasets exhibit consistent improvements of our models over all competitors with statistical significance, demonstrating the effectiveness of our approach. Ablation studies investigate the influence of varying locality and sparsity parameters as well as combinations of multi-scale representations. In future work, we intend to investigate additional instances of the push-based message passing framework, extensions to dynamic graphs and applications to further tasks such as link prediction and graph classification.

\ack This work was done during an internship at CT RDA BAM IBI-US, Siemens Corporate Technology, Princeton, NJ, USA. 

\bibliography{ms}

\end{document}